\documentclass[pmlr,twocolumn,10pt]{jmlr} 

\usepackage{booktabs}
\usepackage{siunitx}

\theorembodyfont{\upshape}
\theoremheaderfont{\scshape}
\theorempostheader{:}
\theoremsep{\newline}

\jmlrvolume{LEAVE UNSET}
\jmlryear{2023}
\jmlrsubmitted{LEAVE UNSET}
\jmlrpublished{LEAVE UNSET}
\jmlrworkshop{Conference on Health, Inference, and Learning (CHIL) 2023} 

\title[Leveraging an Alignment Set in Tackling Instance-Dependent Label Noise]{\hspace{1.1cm} Leveraging an Alignment Set in Tackling \titlebreak Instance-Dependent Label Noise}

\author{%
\Name{Donna Tjandra} \Email{dotjandr@umich.edu}\\
\Name{Jenna Wiens} \Email{wiensj@umich.edu}\\
\addr Computer Science \& Engineering, University of Michigan, Ann Arbor, MI, USA
}

\begin{document}

\newcommand{\anchor}{alignment }
\newcommand{\Anchor}{Alignment }

\maketitle

\begin{abstract}
Noisy training labels can hurt model performance. Most approaches that aim to address label noise assume label noise is independent from the input features. In practice, however, label noise is often feature or \textit{instance-dependent}, and therefore biased (i.e., some instances are more likely to be mislabeled than others). E.g., in clinical care, female patients are more likely to be under-diagnosed for cardiovascular disease compared to male patients. Approaches that ignore this dependence can produce models with poor discriminative performance, and in many healthcare settings, can exacerbate issues around health disparities. In light of these limitations, we propose a two-stage approach to learn in the presence instance-dependent label noise. Our approach utilizes \textit{\anchor points}, a small subset of data for which we know the observed and ground truth labels. On several tasks, our approach leads to consistent improvements over the state-of-the-art in discriminative performance (AUROC) while mitigating bias (area under the equalized odds curve, AUEOC). For example, when predicting acute respiratory failure onset on the MIMIC-III dataset, our approach achieves a harmonic mean (AUROC and AUEOC) of 0.84 (SD [standard deviation] 0.01) while that of the next best baseline is 0.81 (SD 0.01).  Overall, our approach improves accuracy while mitigating potential bias compared to existing approaches in the presence of instance-dependent label noise.
\end{abstract}

\paragraph*{Data and Code Availability}
This paper uses the MIMIC-III dataset \citep{johnson2016mimic}, which is available on the PhysioNet repository \citep{johnson2016physionet}. We also use two public datasets outside of the healthcare domain: 1) the Adult dataset\footnote{https://github.com/AissatouPaye/Fairness-in-Classification-and-Representation-Learning}, and 2) the COMPAS dataset\footnote{https://www.kaggle.com/danofer/compass}. A link to the source code is provided in the footnote\footnote{https://github.com/MLD3/Instance\_Dependent\_Label\_Noise}.

\paragraph*{Institutional Review Board (IRB)}
This work is not regulated as human subjects research since data are de-identified.

\section{Introduction}
\label{sec:intro}

\paragraph{Motivation and Problem Setting} Datasets used to train machine learning models can contain incorrect labels (i.e., label noise), which can lead to overfitting. While label noise is widely studied, the majority of past work focuses on instance-independent label noise (i.e., when the noise is independent from an instance's features) \citep{song2022learning}. However, label noise can depend on instance features \citep{wei2022learning, chang2022disparate}, leading to different noise rates within subsets of the data. Furthermore, in settings where the noise rates differ with respect to a sensitive attribute, this can lead to harmful disparities in model performance \citep{liu2021understanding}. For example, consider the task of predicting cardiovascular disease among patients admitted to a hospital. Compared to male patients, female patients may be more likely to be under-diagnosed \citep{maserejian2009disparities} and thus mislabeled, potentially leading to worse predictions for female patients. Although instance-dependent label noise has recently received more attention \citep{cheng2020learning,xia2020part,wang2021learning}, the effect of these approaches on model bias has been relatively understudied \citep{liu2021understanding}. Here, we address current limitations and propose a novel method for learning with instance-dependent label noise in a setting inspired by healthcare, specifically examining how modeling assumptions affect existing issues around potential model bias.

\paragraph{Gaps in Existing Work} Broadly, current work addressing instance-dependent label noise takes one of two approaches: 1) learn to identify mislabeled instances \citep{cheng2020learningb, xia2022sample, zhu2022detecting}, or 2) learn to optimize a noise-robust objective function \citep{feng2020can, wei2022smooth}. In the first category, instances identified as mislabeled are either filtered out \citep{kim2021fine} or relabeled \citep{berthon2021confidence}. In some settings, this approach can have a negative effect on model bias. Revisiting our example on cardiovascular disease, approaches that filter out mislabeled individuals could ignore more female patients, since they have a potentially higher noise rate. While relabeling approaches use all available data, they can be sensitive to assumptions around the noise distribution \citep{ladouceur2007robustness}. In the second category, current approaches rely on objective functions that are less prone to overfitting to the noise and use all of the data and observed labels \citep{chen2021sln}. However, past work has empirically shown that these improve discriminative performance the most when used to augment filtering approaches, and thus, the limitations and scenarios described above still potentially hold.

\paragraph{Our Idea} In light of these limitations, we propose an approach that addresses instance-dependent label noise, makes no assumptions about the noise distribution, and uses all data during training. We focus on a setting that frequently arises in healthcare, where we are given observed labels for a condition of interest (e.g., cardiovascular disease) and have a clinical expert who can evaluate whether the observed labels are correct for a small subset of the data (e.g., by manual chart review). Using this subset, which we refer to as the `alignment' set, we learn the underlying pattern of label noise in a pre-training step. We then minimize a weighted cross-entropy over all the data. Note that our \anchor set is a special case of anchor points \citep{liu2015classification}, with the added requirement that the set contains instances for which the ground truth and observed labels do \textit{and} do not match.

On synthetic and real data, we demonstrate that our approach improves on state-of-the-art baselines from the noisy labels and fairness literature, such as stochastic label noise \citep{chen2021sln} and group-based peer loss \citep{wang2021fair}. Overall, our contributions include: 
\begin{itemize}
    \item A novel approach to learn from datasets with instance-dependent noise that highlights a setting frequently found in healthcare
    \item A systematic examination of different settings of label noise, evaluating discriminative performance and bias mitigation
    \item Empirical results showing that the proposed approach is robust to both to the noise rate and amount of noise disparity between subgroups, reporting the model’s ability to maintain discriminative performance and mitigate potential bias
    \item A demonstration of how performance of the proposed approach changes when assumptions about the \anchor set are violated
\end{itemize}

\section{Methods}

We introduce a two-stage approach for learning with instance-dependent label noise that leverages a small set of \anchor points for which we have both observed and ground truth labels.

\begin{table}[htbp]
\floatconts
{tab:notation}
{\caption{Notation. A summary of notation used throughout. Superscripts in parentheses specify instances (e.g., $\textbf{x}^{(i)}$). Subscripts specify indexes into a vector (e.g., $\textbf{x}_i$)}}
{
    \resizebox{0.49\textwidth}{!}{ 
    \begin{tabular}{|c|c|}
    \hline
    Notation & Description \\
    \hline \hline
    d & number of features \\
    \hline
    g & number of groups \\
    \hline
    $\textbf{x}\in \mathbb{R}^d$ & feature vector \\
    \hline
    $\hat{y}\in [0, 1]$ & predicted class probabilities \\
    \hline
    $\tilde{y}\in \{-1, 1\}$ & observed label \\
    \hline
    $y\in \{-1, 1\}$ & ground truth label \\
    \hline
    $\hat{\beta}=P(y==\tilde{y} \vert \tilde{y}, \textbf{x}; \phi)$ & prediction of label correctness \\
    \hline
    A & \anchor set, has a instances  \\
    \hline
    $\overline{A}$ & non-\anchor set, $\overline{a}$ instances \\
    \hline
    $\theta$ & main model parameters \\
    \hline 
    $\phi$ & auxiliary model parameters  \\
    \hline
    \end{tabular}}
}
\end{table}

\paragraph*{Notation and Setting} 
Our notation is summarized in \textbf{Table} \textbf{\ref{tab:notation}}, with additional notation defined throughout as needed. Our dataset, $D=A \cup \overline{A}$ consists of instances in $A=\{\textbf{x}^{(j)}, \tilde{y}^{(j)}, y^{(j)}\}_{j=1}^{a}$ and $\overline{A}=\{\textbf{x}^{(i)}, \tilde{y}^{(i)}\}_{i=1}^{\overline{a}}$. $A$ is the set of \anchor points (i.e., the \anchor set), where both $\tilde{y}^{(j)}$ and $y^{(j)}$ are known, and we assume that it includes instances where $\tilde{y}^{(i)} \neq y^{(i)}$. \Anchor points are a special case of anchor points \citep{liu2015classification}, where points that do \textit{and} do not have matching observed and ground truth labels are both required. $\overline{A}$ is the non-\anchor set and contains instances for which we do not know the ground truth labels. In the presence of noisy labels, we assume that whether $\tilde{y}=y$ is dependent on $\textbf{x}$ (i.e., $P(\tilde{y}==y) \neq P(\tilde{y}==y \vert \textbf{x})$). Given this dataset, we aim to train a model to learn $f: \mathbb{R}^d \rightarrow [0, 1]$ (i.e. the function used to predict the ground truth labels), so that we can map unseen instances into one of two classes based on their feature vectors. Our learned model parameters, $\theta$, are such that the output of the corresponding model represents the predicted class probabilities, (i.e., $\hat{y}$). Although we focus on binary classification, our setup can be applied to multiclass classification. 

\paragraph{Justification and Desired Properties} Our setting is inspired by the use of pragmatic labeling tools in healthcare. Such tools are often based on various components of the electronic health record (EHR), and they are applied to identify cohorts or outcomes of interest \citep{upadhyaya2017automated, norton2019development, tjandra2020cohort, yoo2020blood, jain2021visualchexbert}. However, while practical, such definitions are not always reflective of the ground truth, and thus, require validation through manual chart review. This is often done on a randomly chosen subset of individuals, which can be constructed to represent the target population and account for known heterogeneity. As a result, $f$ is the function that predicts whether the condition is actually present, and the \anchor set is the chart reviewed subset used to help learn $f$.

Through our approach, we aim to achieve: 1) robustness to the overall noise rate and 2) robustness to differences in noise rates between groups (i.e., the noise disparity). Revisiting our motivating example with EHR-based labeling tools, previous work has shown that labeling tools for rarer conditions such as drug-induced liver injury and dementia are more likely to be less reliable than those for common conditions \citep{kirby2016phekb}. Similar to how different noise rates can arise in practice, differences in noise rates between subgroups can also vary in practice \citep{kostopoulou2008diagnostic}. As a result, achieving these properties can potentially make our approach generalize to a wide variety of settings. 

\paragraph*{Proposed Approach} Here, we describe the proposed network and training procedure.

\begin{figure*}[h]
    \floatconts
    {fig:proposed_overview}
    {\caption{Our Approach. a)  The model predicts, $\hat{y}$, at training and inference time using $\theta$. At training time, it also predicts whether the observed label is correct using $\phi$. $\theta$ and $\phi$ are pre-trained using $A$ and then fine tuned with the complete dataset. b) We pretrain the model using the \anchor points, then train on the noisy data. $\mathcal{L}_\theta$, $\mathcal{L}_{\phi}$, and $\mathcal{L}_\theta'$ are the objectives for the \anchor points, label confidence score ($\hat{\beta}_{\phi}$), and noisy data, respectively. $\alpha_1$, $\alpha_2$, $\gamma$ are scalar hyperparameters.}}
    {
        \subfigure[Network.]{\label{fig:arch}
             \includegraphics[width=0.45\linewidth]{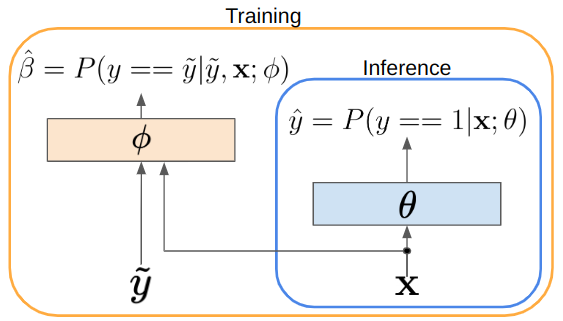} 
            } 
         \qquad       
         \subfigure[Training pipeline.]{\label{fig:pipeline}
             \includegraphics[width=0.47\linewidth]{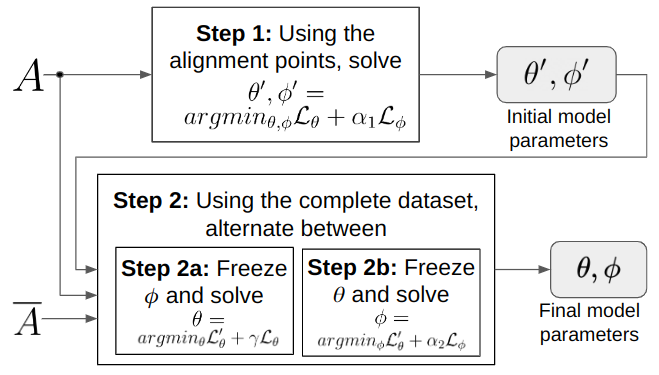}
            }
     }
\end{figure*}

\underline{Proposed Network.} Our proposed network (\textbf{Figure} \textbf{\ref{fig:arch}}) consists of two components. The first, parameterized by $\theta$, is a feed-forward network that uses feature vector $\textbf{x}$ to predict the class probability, $\hat{y}=P(y==1 \vert \textbf{x}; \theta)$. The second component, paramaterized by $\phi$, is an auxiliary feed-forward network that uses observed label $\tilde{y}$ and features $\textbf{x}$ to compute $\hat{\beta}=P(y==\tilde{y} \vert \tilde{y}, \textbf{x}; \phi)$, an instance-dependent prediction for whether the observed label is correct based on $\textbf{x}$ and $\tilde{y}$. $\hat{\beta}$ can be considered as a confidence score for the observed label, with higher values indicating higher confidence. Learning $\hat{\beta}$ models the underlying pattern of label noise by forcing the model to learn which instances are correctly labeled. We use $\hat{\beta}$ to reweight the objective function during the second step of training, as described below. By including the observed label as input to $\phi$, our approach also applies to instance-independent label noise because it accounts for the case when the underlying pattern of label noise does not depend on the features. In order to learn $\hat{\beta}$, we assume that the label noise pattern can be represented as some function, though the specific form of this function (e.g., linear) does not need to be known. During training, we compute the loss using the outputs from both networks. At inference time (i.e., in practical use after training), we compute the class predictions from the network parameterized by $\theta$ only since $\tilde{y}$ is unavailable.

\underline{Training Procedure.} Our training procedure is summarized in \textbf{Figure} \textbf{\ref{fig:pipeline}} and \textbf{Appendix \ref{apd:proposed}}. In Step 1, we pre-train both networks using the \anchor points, $A$, minimizing an objective function based on cross entropy: $\theta', \phi' = argmin_{\theta, \phi} \mathcal{L}_{\theta} + \alpha_1 \mathcal{L}_{\phi}$. $\alpha_1\in\mathbb{R}^+$ is a scalar hyperparameter; $\theta'$ and $\phi'$ are parameters that represent the initial values of $\theta$ and $\phi$. $\mathcal{L}_{\theta}$ is the cross-entropy loss between the class predictions and ground truth labels. It aids in learning the parameter values for $\theta$, and thus, the model's decision boundary. $\mathbb{I}$ is an indicator function. 
\begin{align*}
    &\mathcal{L}_{\theta} = \frac{-1}{\vert A \vert} \sum_{j \in A} 
    \mathbb{I}\left(y^{(j)}==1\right)log\left(\hat{y}^{(j)}\right) \\
    &+ \mathbb{I}\left(y^{(j)}==-1\right)log\left(1-\hat{y}^{(j)}\right)
\end{align*} 
$\mathcal{L}_{\phi}$ is the cross-entropy loss between the predicted confidence score $\hat{\beta}^{(j)}$ and the actual agreement between $\tilde{y}^{(j)}$ and $y^{(j)}$. It aids in learning the weights for $\phi$, and thus, the underlying label noise pattern.
\begin{align*}
    \mathcal{L}_{\phi} &= \frac{-1}{\vert A \vert} \sum_{j \in A}
    \mathbb{I}\left(\tilde{y}^{(j)}==y^{(j)}\right)log \left(\hat{\beta}^{(j)}\right) \\
     &+ \mathbb{I}\left(\tilde{y}^{(j)} \neq y^{(j)}\right)log \left(1 - \hat{\beta}^{(j)}\right)
\end{align*} 
In Step 2, we initialize $\theta$ and $\phi$ as $\theta'$ and $\phi'$ and fine tune using the complete dataset. Step 2 consists of two parts, Step 2a and Step 2b. Each part aims to improve a specific component of the network (e.g., $\theta$) using another component of the network (e.g., $\phi$). We begin with Step 2a, move to Step 2b, and continue to alternate between Step 2a and Step 2b in a manner similar to expectation maximization so that we continually improve both $\theta$ and $\phi$. In Step 2a, we freeze $\phi$ and find $\theta$ that minimizes the objective $\mathcal{L}'_{\theta} + \gamma \mathcal{L}_{\theta}$. $\gamma \in \mathbb{R}^+$ is a scalar hyperparameter. In Step 2b, we freeze $\theta$ and find $\phi$ that minimizes the objective $\mathcal{L}'_{\theta} + \alpha_2 \mathcal{L}_{\phi}$. $\alpha_2 \in \mathbb{R}^+$ is a scalar hyperparameter. $\mathcal{L}_\theta'$ computes the cross-entropy loss over the potentially noisy, non-\anchor points. Each instance is weighted by the model's confidence in whether the observed label is correct via $\hat{\beta}^{(i)}$, taking advantage of the model's learned noise pattern. Our approach aims to mitigate bias by up-weighting groups, $k=1,2,...,g$ with a higher estimated noise rate, $\hat{r}_k$, so that they are not dominated by/ignored compared to groups with a lower estimated noise rate. 
\begin{align*} 
    &\mathcal{L}_\theta' = \frac{-1}{\vert \overline{A} \vert} \sum_{k=1}^{g} \frac{1}{1-\hat{r}_k} \sum_{i \in \overline{A} \cap G_k} \\  &\sum_{j\in\{-1, 1\}}\hat{\beta}^{(i)}_{\phi}\mathbb{I}\left(\tilde{y}^{(i)}==j\right)log\left(\hat{y}^{(i)}_j\right)  
\end{align*}
We calculate $1 - \hat{r}_k$ is as follows. We introduce sets $G_k$ for $k=1,2,...,g$ to represent disjoint subgroups of interest in the data, which are assumed to be known in advance. $G_a \cap G_b = \emptyset$ for all $a=1, 2, ..., g$, $b=1, 2, ..., g$ with $a \neq b$ and $\cup_{k=1}^{g} G_k = D$. Each group $G_k$ is then associated with estimated noise rate $\hat{r}_k=\frac{1}{\vert G_k \vert} \sum_{i \in G_k} 1-\hat{\beta}^{(i)}$. Although weighting each instance by $\hat{\beta}$ is a form of soft filtering, weighting each group by the inverse of its overall `clean' rate avoids the effect of de-emphasizing groups with higher predicted noise rates. As a result, the expected value of $\mathcal{L}_\theta'$ with respect to $\hat{\beta}$ is equal to the cross-entropy loss between the model's predictions and ground truth labels (see \textbf{Appendix \ref{apd:proposed}} for proof). However, this assumes accurate estimates of $\hat{\beta}$. Thus, we expect that the proposed approach will perform best when the \anchor set is representative of the target population. In scenarios where the \anchor set is biased (e.g., some groups are underrepresented), if the learned noise function does not transfer to the underrepresented group, then the proposed approach may not be beneficial. In \textbf{Section \ref{sec:res}}, we test this. 

During Step 2a, $\mathcal{L}_\theta'$ is used to train $\theta$ by learning to predict $\hat{y}$ such that it matches observed label $\tilde{y}$ on instances that are predicted to be correctly labeled. During Step 2b, $\mathcal{L}_\theta'$ is used to train $\phi$. Here, since $\theta$ is frozen and $\phi$ is not, the network learns to predict the optimal $\hat{\beta}$. Based on $\mathcal{L}_\theta'$ alone, there are two possible options to learn $\hat{\beta}$: 1) consistently make $\hat{\beta}$ close to 0, and 2) predict $\hat{\beta}$ such that it is close to 1 when $\hat{y}$ matches $\tilde{y}$ and close to 0 when $\hat{y}$ does not match $\tilde{y}$. Since $\tilde{y}$ is used as a proxy for $y$ in this step, the second option aligns with what we want $\hat{\beta}$ to represent. To encourage this over the first option (i.e., consistently predicting 0 for $\hat{\beta}$), we include $\mathcal{L}_{\phi}$ in Step 2b, which is \textit{not} minimized by consistently predicting 0 for $\hat{\beta}$. Note that, in Step 2b, we rely on the cluster assumption \citep{singh2008unlabeled} from semi-supervised learning, which broadly states that labeled data fall into clusters and that unlabeled data aid in defining these clusters. In the context of Step 2b, `labeled' and `unlabeled' are analogous to whether we know if the ground truth and observed labels match (i.e., \anchor point versus non-\anchor point), rather than the actual class labels themselves. As a result, we also rely on the \anchor set being representative of the target population here to avoid dataset shift.

In contrast to previous filtering approaches, our approach utilizes all data during training. Moreover, it does not require a specialized architecture beyond the auxiliary network to compute $\hat{\beta}$. Thus, it can be used to augment existing architectures.

\section{Experimental Setup}

We empirically explore the performance of our proposed approach relative to state-of-the-art baselines on five benchmark prediction tasks with two different label noise settings. For reproducibility, full implementation details are provided in \textbf{Appendices \ref{apd:preprocess}} and \textbf{\ref{apd:net_and_train}}. We aim to test 1) the extent to which our desired properties hold, 2) the extent to which the proposed approach is robust to changes in the composition of the \anchor set, and 3) which components of the proposed approach contribute the most.

\paragraph*{Datasets} We consider five different binary prediction tasks on four datasets from several domains with synthetic and real datasets. Though inspired by healthcare, we also consider domains outside of healthcare to show the broader applicability of our approach in areas where harmful biases can arise (e.g., predicting recidivism and income). Throughout our experiments, we start by assuming the labels in the dataset are noise free, and we inject varying amounts of synthetic label noise. In this subsection, we describe the tasks, features, and `ground truth' labels we use. The next subsection will describe how we introduce synthetic label noise.

\underline{Synthetic}: We generate a dataset containing 5,000 instances according to the generative process in \textbf{Appendix \ref{apd:preprocess}}. The positive rates for the majority and minority groups are 37.5\% and 32.3\%, respectively. 

\underline{MIMIC-III}: Within the healthcare domain, we leverage a publicly available dataset of electronic health record data \citep{johnson2016mimic}. We consider two separate prediction tasks: onset of 1) acute respiratory failure (ARF) and 2) shock in the ICU (intensive care unit) \citep{Oh2019RelaxedPS}. MIMIC-III includes data pertaining to vital signs, medications, diagnostic and procedure codes, and laboratory measurements. We consider the four hour prediction setup for both tasks as described by \cite{tang2020democratizing}, resulting in 15,873 and 19,342 ICU encounters, respectively. After preprocessing (see \textbf{Appendix \ref{apd:preprocess}}), each encounter had 16,278 and 18,186 features for each task respectively. We use race as a sensitive attribute, with about 70\% of patients being white (positive rate 4.5\% [ARF], 4.1\% [shock]) and 30\% being non-white (positive rate 4.4\% [ARF], 3.7\% [shock]).

\ 

\noindent Beyond healthcare, we use two benchmark datasets frequently considered in the fairness domain.

\underline{Adult}: a publicly available dataset of census data \citep{adultDataset}. We consider the task of predicting whether an individual's income is over \$50,000. This dataset includes data pertaining to age, education, work type, work sector, race, sex, marital status, and country. Its training and test sets contain 32,561 and 16,281 individuals, respectively. We use a pre-processed version of this dataset and randomly select 1,000 individuals out of 32,561 for training. We also only include features pertaining to age, education, work type, marital status, work sector, and sex to make the task more difficult (see \textbf{Appendix \ref{apd:preprocess}}). After preprocessing, each individual was associated with 56 features, and all features had a range of 0-1. We use sex as a sensitive attribute, with 67.5\% of individuals being male (positive rate 30.9\%) and 32.5\% being female (positive rate 11.3\%).

\underline{COMPAS}: a publicly available dataset collected by ProPublica from Broward County, Florida, USA \citep{compasArticle}. We consider the task of predicting recidivism within two years, i.e., whether a criminal defendant is likely to re-offend. COMPAS includes data pertaining to age, race, sex, and criminal history. We use a pre-processed version of this dataset and also normalize each feature to have a range of 0-1 (see \textbf{Appendix \ref{apd:preprocess}}). After preprocessing, the dataset included 6,172 individuals with 11 features per individual. We use race as a sensitive attribute, with 65.8\% of individuals being white (positive rate 39.1\%) and 34.2\% being non-white (positive rate 44.5\%).

\paragraph*{Label Noise} To test the robustness of our approach in different settings of label noise, we introduce synthetic instance-dependent label noise to our datasets. Like past work \citep{song2022learning}, our setup is limited for the real datasets because our added noise is synthetic and we use the labels provided in the dataset as ground truth, since we do not have access to actual ground truth labels on these public datasets. 

To introduce instance-dependent noise, mislabeling was a function of the features. Let $\textbf{w}_m \sim N(0, 0.33)^{D}$ and $z_m = \sigma(\textbf{x} \cdot \textbf{w}_m)$, where $\sigma$ is the sigmoid function, denote the coefficients describing the contribution of each feature to mislabeling and the risk of mislabeling, respectively. Whether an instance was mislabeled was based on $z_m$ and the desired noise rate. For example, for a noise rate of 30\%, instances whose value for $z_m$ was above the $70^{th}$ percentile had their labels flipped. This allowed us to vary the noise rate within subgroups in a straightforward manner. Across datasets, we focused on cases where the noise rate in the `minority' population was always greater than or equal to that of the `majority' group since this is more likely to occur \citep{suite2007beyond}.

\paragraph*{Evaluation Metrics} We evaluate our proposed approach in terms of discriminative performance and model bias. For discriminative performance, we evaluate using the area under the receiver operating characteristic curve (AUROC) (higher is better). 

With respect to model bias, while there exist many different measures, we focus on equalized odds \citep{hardt2016equality}, since it is commonly used in the context of healthcare \citep{pfohl2019counterfactual, xu2022algorithmic, yogarajan2023data}, when similar performance across groups is desired \citep{rajkomar2018ensuring, pfohl2021empirical}. Because equalized odds focuses on the difference between the true and false positive rates among groups, it is applicable to many settings in healthcare since the consequences of failing to treat a patient in need \citep{pingleton1988complications, bone1994sepsis}, or giving an inappropriate treatment \citep{bogun2004misdiagnosis, nasrallah2015consequences} can be serious. More specifically, we measure the area under the equalized odds curve (AUEOC) \citep{de2020fairness} (higher is better). For classification threshold $\tau$, we calculate the equalized odds (EO($\tau$)) between two groups, called 1 and 2, as shown below. $TP_a(\tau)$ and $FP_a(\tau)$ denote true and false positive rates for group $a$ at threshold $\tau$, respectively. The AUEOC is obtained by plotting the EO against all possible values of $\tau$ and calculating the area under the curve. 

We compute the harmonic mean (HM) between the AUROC and AUEOC to highlight how the different approaches simultaneously maintain discriminative performance and mitigate bias. In the harmonic mean the worse performing metric dominates. For example, if a classifier has AUROC=0.5 and AUEOC=1.0, the harmonic mean will emphasize the poor discriminative performance.
 
\begin{align*}
    EO(\tau) = \frac{2 - \vert TP_1(\tau) - TP_2(\tau) \vert - \vert FP_1(\tau) - FP_2(\tau) \vert }{2}
\end{align*}

\paragraph*{Baselines} We evaluate our proposed approach with several baselines to test different hypotheses. 

\underline{Standard} does not account for label noise and assumes that $\tilde{y}=y$ is always true. 

\underline{SLN + Filter} \citep{chen2021sln} combines filtering \citep{arpit2017closer} and SLN \citep{chen2021sln} and was shown to outperform state-of-the-art approaches like Co-Teaching \citep{han2018co} and DivideMix \citep{li2019dividemix}. It relies on filtering heuristics, which indirectly rely on uniform random label noise to maintain discriminative performance and mitigate bias. 

\underline{JS (Jensen-Shannon) Loss} \citep{englesson2021generalized} builds on semi-supervised learning and encourages model consistency when predicting on perturbations of the input features. It was shown to be competitive with other state-of-the-art noise-robust loss functions \citep{ma2020normalized}. It was proposed for instance-independent label noise. 

\underline{Transition} \citep{xia2020part} learns to correct for noisy labels by learning a transition function and was shown to outperform state-of-the-art approaches such as MentorNet \citep{jiang2018mentornet}. It applies to instance-dependent label noise, but it assumes that the contributions of each feature to mislabeling and input reconstruction are identical. 

\underline{CSIDN} (confidence-scored instance-dependent noise) \citep{berthon2021confidence} also learns a transition function and was shown to outperform state-of-the-art approaches such as forward correction \citep{patrini2017making}. Like our approach, CSIDN uses the concept of `confidence' in the observed label to help with training. Unlike our approach, CSIDN uses the model's class predictions directly as confidence scores (instead predicting them via an auxiliary network) and uses them to learn the transition function (as opposed to re-weighting the loss). 

\underline{Fair GPL} \citep{wang2021fair} builds on work addressing uniform random label noise \citep{jiang2020identifying} and uses peer loss (i.e., data augmentation that reduces the correlation between the observed label and model's predictions) within subgroups \citep{wang2021fair}. It assumes that label noise only depends on group membership. 

We also train a model using the ground truth labels (called \underline{Clean Labels}) as an empirical upper bound for discriminative performance.

\paragraph*{Implementation Details}
For each dataset, we randomly split the data into 80/20\% training/test, ensuring that data from the same individual did not appear across splits. For the Adult dataset, we used the test set provided and randomly selected 1,000 individuals from the training set. We then randomly selected 10\% of the training data for all datasets except MIMIC-III from each subgroup to be \anchor points, thereby ensuring that they were representative of the overall population. For the MIMIC-III dataset, 2\% from each subgroup were selected as \anchor points due to the larger size of the dataset. \Anchor points were selected randomly to simulate our setting of focus, where we have a proxy labeling function and then randomly select a subset of the data to chart review in order to validate the proxy function. Then, for all datasets, half of the \anchor points were then set aside as a validation set to use during training for early stopping and hyperparameter selection, while the other half remained in the training set. Later, in our experiments, we evaluated when the \anchor set size varied and when the \anchor set was biased. All approaches (i.e., baselines and proposed) were given the ground truth labels for data in the \anchor set (i.e., no noise added to \anchor points) during training so that some approaches did not have an unfair advantage.

All models were trained in Python3.7 and Pytorch1.7.1 \citep{paszke2017automatic}, using Adam \citep{kingma2014adam}. Hyperparameters, including the learning rate, L2 regularization constant, and objective function scalars (e.g., $\alpha$), were tuned using random search, with a budget of 20. We used early stopping (patience=10) based on validation set performance, which we measured with the HM. We report results on the held-out test set, showing the mean and standard deviation over 10 replications.

\section{Results and Discussion}
\label{sec:res}
 
We describe the results from experiments with instance-dependent noise. For each plot, we combined discriminative performance and bias mitigation and plotted the HM of the AUROC and AUEOC to assess general performance with respect to both metrics. We show the AUROC and AUEOC separately in \textbf{Appendix \ref{apd:extra_exp}}. Additional experiments are provided in \textbf{Appendix \ref{apd:extra_exp}}. Their results are summarized here.

\paragraph*{Robustness to Noise Rate} Here, we investigated how robust the proposed approach and baselines were to varying amounts of instance-dependent label noise (\textbf{Figure} \textbf{\ref{fig:res_noise_exp}}). Since noise was synthetically introduced and not dataset specific, we conducted two experiments on the synthetic dataset. In the first, we varied the overall noise rate from 10-60\% in the majority group. For the minority group, we considered noise rates that were consistently 20\% higher than that of the majority group, to keep the noise disparity level (i.e., the difference in noise rates between subgroups) constant. In the second, we varied the minority noise rate from 20-90\% with a majority noise rate fixed at 20\% throughout (i.e., from 0-70\% disparity) on the synthetic dataset. 

\begin{figure*}[h]
    \floatconts
    {fig:res_noise_exp}
    {\caption{Robustness to noise rate and noise disparity in an instance dependent setting. We plot the mean and standard deviation for 10 random seeds. As the noise rate (a) and disparity (b) increase, the proposed approach generally shows the least degradation up to a minority noise rate of 80\%.}}
    {
        \subfigure[Noise rate.]{\label{fig:res_noise_rate}
             \includegraphics[width=0.4\linewidth]{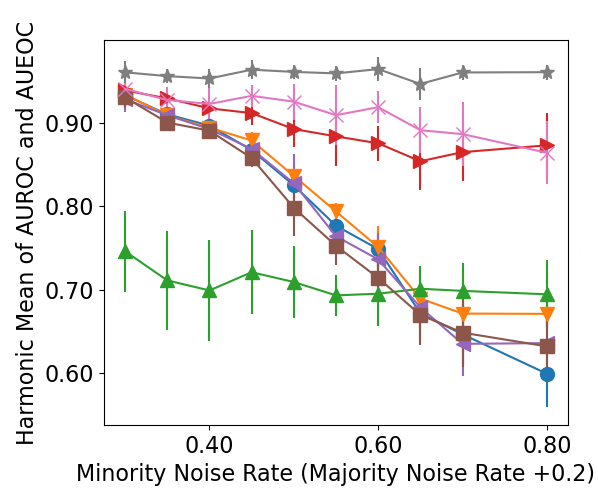}  
            } 
         \qquad       
         \subfigure[Noise disparity. ]{\label{fig:res_noise_disparity}
             \includegraphics[width=0.52\linewidth]{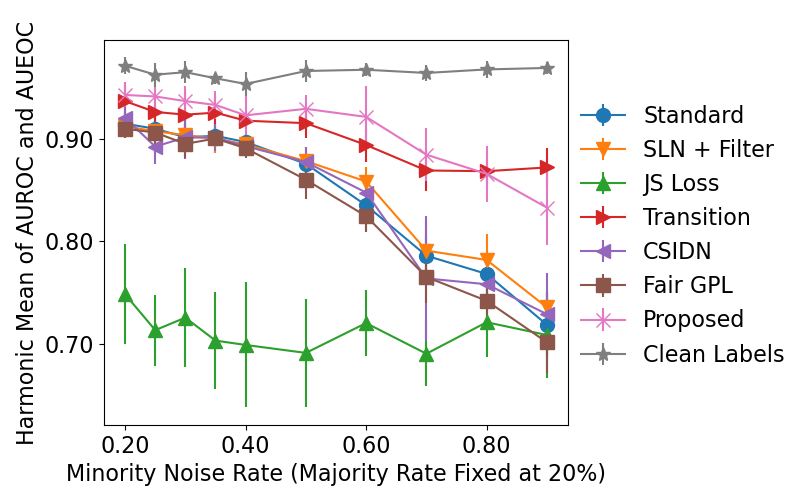} 
            }
     }
\end{figure*}

\underline{Part 1: Overall Noise Rate.} Overall, our proposed approach demonstrated robustness to a variety of noise rates within a realistic range (\textbf{Figure} \textbf{\ref{fig:res_noise_rate}}). At low minority noise rates (i.e., below 40\%), the proposed approach and baselines, with the exception of JS Loss, were competitive. As the noise rate increased, many of the baselines experienced noticeable degradation in performance. The proposed approach and Transition showed more robustness, with the proposed approach being the most robust until a minority noise rate of 80\%, which represents an extreme case of label noise. 

\underline{Part 2: Noise Disparity.} Like the previous experiment, the proposed approach was robust over a variety of noise disparities (\textbf{Figure} \textbf{\ref{fig:res_noise_disparity}}). This is likely because the objective function $\mathcal{L}'_\theta$ from Step 2 of training accounts for disparities by scaling each instance-specific loss term with the reciprocal of its estimated group clean rate (i.e., 1 - the estimated group noise rate). Similar to the previous experiment, at a minority noise rate of 80\% and above, the proposed approach was no longer the most robust, though this setting is unlikely to occur in practice.

\begin{figure*}[h]
    \floatconts
    {fig:res_anchor_exp}
    {\caption{Robustness to varying \anchor sets. Mean and standard deviation for 10 random seeds.}}
    {
        \subfigure[As we decrease the \anchor set size (proportion of training data) performance decreases. Still, at an \anchor set size of 3\%, the proposed approach is robust.]{\label{fig:res_asize}
             \includegraphics[width=1\linewidth]{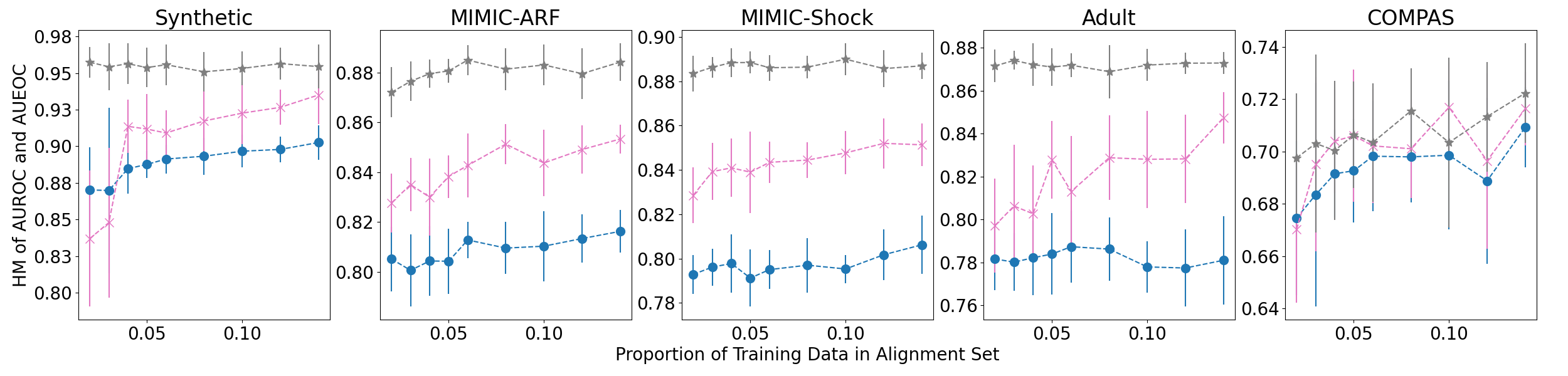} 
            } 
         \qquad       
         \subfigure[We varied the \anchor set bias (proportion of minority instances). The proposed approach is generally robust to variations in bias. The dashed vertical black line shows an unbiased \anchor set.]{\label{fig:res_adistr}
             \includegraphics[width=1\linewidth]{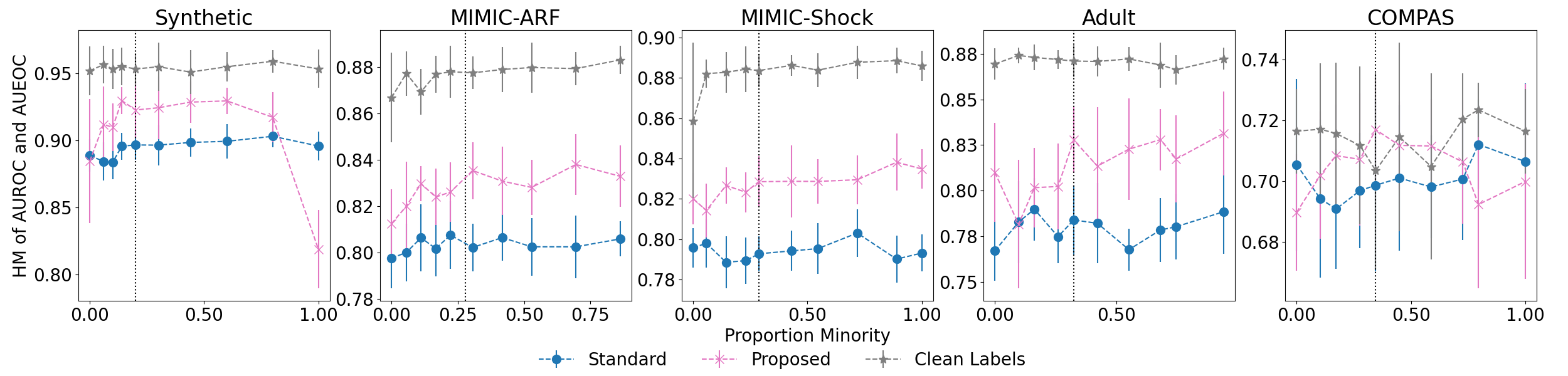} 
            }
     }
\end{figure*}

\paragraph*{Sensitivity to \Anchor Set Composition} Our next set of experiments tested the proposed approach in settings where we relax key settings about the \anchor set. We considered all datasets with instance-dependent noise. The majority/minority noise rates were 20\%/40\%, respectively. Here we show performance with respect to the proposed approach, Standard, and Clean Labels. Results for the other baselines are included in \textbf{Appendix \ref{apd:extra_exp}}.

\underline{Part 1: \Anchor set size}. We varied the size of the \anchor set, from 1\% and 15\% of the training set, with the \anchor set being representative of the test set (\textbf{Figure} \textbf{\ref{fig:res_asize}}). The proposed approach was robust to a wide range of \anchor set sizes, only showing noticeable degradation at \anchor set sizes of 3\% or lower. As the size of the \anchor set grew, performance improved, likely since having a larger \anchor set provided access to a larger set of ground truth labels at training time. Although the minimum number of points required in the \anchor set is likely to vary depending on the task, our results are promising in that they show that our approach is effective on a variety of real life tasks, even when the \anchor set is small (i.e., as little as 3\% of the data).

\underline{Part 2: Biased \anchor set}. Here, we test how the proposed approach performs when the \anchor set is not representative of the population. We varied the amount of bias in the \anchor set by changing the proportion at which the subgroups were present. We kept the size of the \anchor set constant at 10\% of the training data (2\% for MIMIC-III on both tasks). We observed that the proposed approach was robust over a wide range of conditions, i.e., when the minority proportion is 20\%-80\% (\textbf{Figure} \textbf{\ref{fig:res_adistr}}). We hypothesize that this is because the learned relationship between the features and noise can generalize across groups to an extent. In scenarios where performance of the proposed approach degraded, one subgroup heavily dominated the \anchor set. This is shown in \textbf{Figure} \textbf{\ref{fig:res_adistr}} on the extremes of the x-axis of some datasets, which correspond to an \anchor set that is heavily over-represented for one subgroup and heavily under-represented for the other. Our approach relies, in part, on having a relatively unbiased \anchor set for estimating $\hat{\beta}$ in order to avoid introducing dataset shift between the two steps of our training pipeline. Thus, these results are in line with our expectations and highlight a limitation of our approach. However, despite this reliance, we observe that our approach is still robust in some scenarios where the \anchor set is biased. 

\paragraph*{Which Parts of Our Approach Matter?} Our last set of results examines the individual components of the approach itself on the synthetic dataset. Here, we performed an ablation study where we began with training on only the \anchor points (i.e., Step 1 of our approach), and then gradually added the other components of our approach (e.g., add Step 2a). In summary, while each component improved performance, we find that the most improvement came from adding $\mathcal{L}_{\theta}$ and $\mathcal{L}_{\phi}$ during Steps 2a and 2b, respectively, as opposed to using only $\mathcal{L}_{\theta}'$ during those steps. We also performed a hyperparameter sensitivity analysis on the three hyperparameters, $\alpha_1$, $\gamma$, and $\alpha_2$, that our approach introduced. The approach was most sensitive to the $\alpha_2$ hyperparameter and more robust to $\alpha_1$ and $\gamma$. We include results for the ablation study and hyperparameter sensitivity analysis in \textbf{Appendix \ref{apd:extra_exp}}.

\paragraph*{Which Parts of Our Approach Matter?} Our last set of results aims to more closely examine the individual components of the approach itself. We include results for an ablation study and a hyperparameter sensitivity analysis in \textbf{Appendix \ref{apd:extra_exp}}. In summary, while each component improved performance, we find that the most improvement came from adding $\mathcal{L}_{\theta}$ and $\mathcal{L}_{\phi}$ during Steps 2a and 2b, respectively, as opposed to using only $\mathcal{L}_{\theta}'$ during those steps. The approach was most sensitive to the $\alpha_2$ hyperparameter and more robust to $\alpha_1$ and $\gamma$.

\section{Related Work}

We build from previous work in label noise and address key limitations. Generally, many state-of-the-art approaches \citep{song2022learning} are limited in that they do not consider instance-dependent noise, do not consider the potential consequences of bias in label noise, or do not leverage the information our setting provides. We tackle these limitations by accounting for differences in noise rates among subsets of the data and taking advantage of additional information that can be found in our setting. In this section, we summarize past work and highlight our contributions.

\paragraph*{Identifying Mislabeled Data} Approaches that learn to identify mislabeled instances fall into two sub-categories: 1) filtering approaches and 2) relabeling approaches. Filtering approaches use heuristics to identify mislabeled instances (e.g., MentorNet \citep{jiang2018mentornet}, Co-teaching \citep{han2018co}, FINE \citep{kim2021fine}). Many are based on the idea that correctly labeled instances are easier to classify than mislabeled instances (i.e., the memorization effect) \citep{arpit2017closer}. For example, mislabeled instances could be those that the model incorrectly classifies \citep{verbaeten2002identifying, khoshgoftaar2004generating, thongkam2008support, chen2019understanding}, have a high loss value \citep{yao2020searching}, or significantly increase the complexity of the model \citep{gamberger1996noise}. Given the identified mislabeled instances, these approaches either ignore them during training \citep{zhang2020self} or treat them as `unlabeled’ and apply techniques from semi-supervised learning (e.g., DivideMix \citep{li2019dividemix}, SELF \citep{nguyen2019self}). Overall, these heuristics have been shown to improve discriminative performance. However, depending on the setting, they can disproportionately discard subsets of data, which could exacerbate biases in model performance.

For binary classification, some approaches `correct' (i.e., switch) the observed label for instances that are predicted to be incorrect \citep{han2020sigua, zheng2020error}. Building on this idea, others make use of a transition function that estimates the probability of the observed label being correct. Model predictions can then be adjusted by applying the transition function to the classifier's predictions for each class. Some works manually construct the transition function from expert knowledge \citep{patrini2017making}, while others learn it \citep{xiao2015learning, xu2019l_dmi, yao2020dual, zheng2021meta, jiang2022information, bae2022noisy, chengclass, li2022estimating}. However, such approaches often make assumptions on the form of the noise distribution, and past work has shown that results are sensitive to the choice of distribution \citep{ladouceur2007robustness}.

To date, much of the work described above assumes instance-independent label noise (i.e., mislabeling is independent of the features). However, when this assumption is violated, the model may overfit to label noise \citep{lukasik2020does}. From an emerging body of work in instance-dependent label noise \citep{cheng2020learning, xia2020part, wang2021tackling, zhu2022beyond}, current approaches remain limited in that they still rely on filtering heuristics. Although we use soft filtering, we filter based on the learned relationship between the features and noise rather than existing heuristics and upweight groups with a higher estimated noise rate. While similar to a transition function in some aspects, our approach requires fewer probability estimates on label correctness (two estimates compared to the number of classes squared for a transition function) while achieving state-of-the-art performance.

\paragraph*{Noise-Robust Loss Functions} Prior work examines how regularization techniques can be adapted to the noisy labels setting, addressing issues related to overfitting on noisy data \citep{menon2019can, lukasik2020does, englesson2021generalized}. Label smoothing, and in some cases negative label smoothing, were found to improve the accuracy on both correctly labeled and mislabeled data \citep{lukasik2020does, wei2022smooth}. With this approach, the observed labels are perturbed by a small, pre-determined value, with all labels receiving the same perturbation at every training epoch. Follow-up work found that, instead of applying the same perturbation at each epoch, adding a small amount of Gaussian stochastic label noise (SLN) at each epoch resulted in further improvements, as it helped to escape from local optima \citep{chen2021sln}. However, these approaches were most beneficial in the context of augmenting existing methods that identify mislabeled instances (e.g., stochastic label noise is applied to instances that are identified as correctly labeled by filtering approaches), and thus, potentially suffer from the same limitations. Alternatively, recent work has also proposed perturbing the features to encourage consistency in the model's predictions \citep{englesson2021generalized}, though mainly in the context of instance-independent label noise. Others have proposed noise-robust variations of cross entropy loss \citep{feng2020can, wang2021learning} but generally relied on assumptions like the memorization effect.

\paragraph*{Label Noise in Fairness} Label noise has also been addressed within the fairness literature recently. When the frequencies at which subgroups (defined by a sensitive attribute) appear are different within a dataset, past work has shown that common approaches addressing label noise can increase the prediction error for minority groups (i.e., rarer subgroups) \citep{liu2021understanding}. Past work proposed to re-weight instances from subgroups during training where model performance is poorer \citep{jiang2020identifying} in the instance-independent noise setting. Others use peer loss \citep{liu2020peer} within subgroups \citep{wang2021fair} but assume that noise depends only on the sensitive attribute. We also train with a weighted loss, but weights are based on predicted label correctness rather than performance on the observed labels. Recently, \cite{wu2022fair} addressed some of the gaps of past work by examining the instance-dependent case. Our proposed approach differs from theirs in that we do not require our features to be grouped into distinct categories, such as root and low level attributes.

\paragraph*{Anchor Points for Addressing Label Noise} Another related setting in past work uses anchor points. Anchor points are subsets of the data where the ground truth labels are known \citep{liu2015classification}. To date, anchor points are generally used to learn a transition function \citep{xia2019anchor, xia2020part, berthon2021confidence} or for label correction directly \citep{wu2021learning}. We use a similar concept, \anchor points, to 1) pre-train the model, and 2) predict label correctness. The first part builds from work in semi-supervised learning \citep{cascante2021curriculum}, which has shown improvements from pre-training on labeled data. The second part is similar to a transition function, but differs in that we use the correctness predictions to re-weight the loss rather than adjust the predictions. We also assume that, for some alignment points, the ground truth and observed labels do not match. Generally, anchor-based approaches mitigate model bias by implicitly assuming that the anchor points are representative of the target population. Our approach also uses this assumption, but we empirically explore how model performance changes when the anchor points are biased (i.e., not representative), since it may be easier to obtain correct labels for specific subgroups \cite{spector2021respecting}.

\section{Conclusion}

We introduce a novel approach for learning with instance-dependent label noise. Our two-stage approach uses the complete dataset and learns the relationship between the features and label noise using a small set of \anchor points. On several datasets, we show that the proposed approach leads to improvements over state-of-the-art baselines in maintaining discriminative performance and mitigating bias. Our approach is not without limitations. We demonstrated that the success of the approach depends, in part, on the representativeness in the \anchor set. Our experiments were also on pseudo-synthetic data in which we injected noise; this assumes we start from a noise free dataset. Finally, we only examined one form of bias in a specific case of instance-dependent label noise. Nonetheless, our case frequently arises in healthcare, especially when pragmatic (e.g., automated) labeling tools are used on large datasets, and chart review on the entire dataset is infeasible.


\acks{This work was supported by Cisco Research and the National Science Foundation (NSF award no. IIS 2124127). The views and conclusions in this document are those of the authors and should not be interpreted as necessarily representing the official policies, either expressed or implied, of Cisco Systems Inc. or the National Science Foundation. We also thank the anonymous reviewers for their valuable feedback.}  

\bibliography{sources.bib}

\clearpage

\appendix

\section{Proposed Approach: Additional Details} \label{apd:proposed}

We provide additional details on our approach, including a general overview in the form of pseudocode as well as a justification for the proposed objective function and its relation to the clean label loss.

\subsection{General Overview}

We summarize our approach with pseudocode below in \textbf{Algorithm \ref{ag:proposed}}. We begin with the dataset and initial model parameters, and we aim to use the dataset to learn the final model parameters. A is the set of anchor points. $\theta'$ and $\phi'$ are the initial model parameters for the $\theta$ and $\phi$ networks. Here, 'stopping criteria' may refer to any stopping criteria, such as early stopping. The Freeze() function takes as input model parameters and freezes them, and the Unfreeze() function takes as input model parameters and unfreezes them.

\begin{algorithm}[htbp]
\floatconts
    {ag:proposed}
    {\caption{Proposed approach.}}
    {
        \KwIn{\{$\textbf{x}^{(i)}, \tilde{y}^{(i)}, y^{(i)}\}_{i \in A}$, \{$\textbf{x}^{(i)}, \tilde{y}^{(i)}\}_{i \notin A}$, $\theta', \phi'$ }  
        \KwOut{$\theta, \phi$ (final model parameters)} 
        \textbf{Hyperparameters:} Scalars $\alpha_1, \alpha_2, \gamma$ 

        Train (\{$\textbf{x}^{(i)}, \tilde{y}^{(i)}, y^{(i)}\}_{i \in A}$, \{$\textbf{x}^{(i)}, \tilde{y}^{(i)}\}_{i \notin A}$, $\theta', \phi'$) 
        \begin{enumerate}
            \item While $\neg$(stopping criteria) (Step 1) 
            \begin{enumerate}
                \item $\hat{y}=\theta'(\textbf{x})$ (Predict label) 
                \item $\hat{\beta}_{\phi}=\phi'(\textbf{x}, \tilde{y})$ (Predict label confidence) 
                \item $\mathcal{L}_{\theta} = \frac{-1}{\vert A \vert} \sum_{i \in A} \sum_{j=1}^{c}  \mathbb{I}\left(y^{(i)}==j\right)log\left(\hat{y}^{(i)}_j\right)$  
                \item $\mathcal{L}_{\phi} = \frac{-1}{\vert A \vert} \sum_{i \in A} \mathbb{I}\left(\tilde{y}^{(i)}==y^{(i)}\right)log \left(\hat{\beta}^{(i)}_{\phi}\right) + \mathbb{I}\left(\tilde{y}^{(i)} \neq y^{(i)}\right)log \left(1 - \hat{\beta}^{(i)}_{\phi}\right)$ 
                \item Loss = $\mathcal{L}_{\theta} + \alpha_1 \mathcal{L}_{\phi}$ 
                \item Update model parameters 
                \item Compute stopping criteria 
            \end{enumerate}
            \item $\theta, \phi \gets \theta', \phi'$
            \item Freeze($\phi$)
            \item While $\neg$(stopping criteria) (Step 2)
                \begin{enumerate}
                    \item $\hat{y}=\theta'(\textbf{x})$ 
                    \item $\hat{\beta}_{\phi}=\phi'(\textbf{x}, \tilde{y})$ 
                    \item $\mathcal{L}_{\theta} = \frac{-1}{\vert A \vert} \sum_{i \in A} \sum_{j=1}^{c} \mathbb{I}\left(y^{(i)}==j\right)log\left(\hat{y}^{(i)}_j\right)$ 
                    \item $\mathcal{L}_{\phi} = \frac{-1}{\vert A \vert} \sum_{i \in A} \mathbb{I}\left(\tilde{y}^{(i)}==y^{(i)}\right)log \left(\hat{\beta}^{(i)}_{\phi}\right) + \mathbb{I}\left(\tilde{y}^{(i)} \neq y^{(i)}\right)log \left(1 - \hat{\beta}^{(i)}_{\phi}\right)$ 
                    \item $\mathcal{L}_\theta' = \frac{-1}{\vert \overline{A} \vert} \sum_{k=1}^{G} \frac{1}{1-\hat{r}_k} \sum_{i \in \overline{A} \cap g_k} \sum_{j=1}^{c} \\ \hat{\beta}^{(i)}_{\phi}\mathbb{I}\left(\tilde{y}^{(i)}==j\right)log\left(\hat{y}^{(i)}_j\right)$ (Weighted loss) 
                    \item If {$\phi$ is frozen} (Step 2a) 
                        \begin{enumerate}
                        \item Loss = $\mathcal{L}_{\theta'} + \gamma \mathcal{L}_{\theta}$ 
                        \item Unfreeze($\phi$) 
                        \item Freeze($\theta$) 
                        \end{enumerate}
                    \item Else (Step 2b) 
                        \begin{enumerate}
                        \item Loss = $\mathcal{L}_{\theta'} + \alpha_2 \mathcal{L}_{\phi}$ 
                        \item Unfreeze($\theta$) 
                        \item Freeze($\phi$) 
                        \end{enumerate}
                    \item Update model parameters 
                    \item Compute stopping criteria 
                \end{enumerate}
            \item Return $\theta, \phi$ (Final model parameters)
        \end{enumerate}
    }
\end{algorithm}

\subsection{Proposed and Clean Label Loss}

We show that minimizing the proposed loss $\mathcal{L}'_\theta$ from Step 2 of the proposed method is equal to minimizing cross entropy on the clean labels in expectation. 
\begin{align*}
    \mathcal{L}'_\theta &= \frac{-1}{\vert \overline{A} \vert} \sum_{k=1}^{g} \sum_{i \in \overline{A} \cap G_k} \frac{1}{1-\hat{r}_k} \sum_{j=1}^{c} \\ 
    & \hat{\beta}^{(i)}_{\phi}\mathbb{I}\left(\tilde{y}^{(i)}==j\right)log\left(\hat{y}^{(i)}_j\right)   \\
\end{align*}
Therefore,
\begin{align*} 
    &\mathbb{E} \left[\sum_{k=1}^{g} \sum_{i \in \overline{A} \cap G_k} \frac{1}{1-\hat{r}_k} \sum_{j=1}^{c}\hat{\beta}^{(i)}_{\phi}\mathbb{I}\left(\tilde{y}^{(i)}==j\right)log\left(\hat{y}^{(i)}_j\right) \right] \\ 
    & = \sum_{k=1}^{g} \sum_{i \in \overline{A} \cap G_k} \frac{1}{1-\hat{r}_k} \sum_{j=1}^{c}\mathbb{E} \left [\hat{\beta}_{\phi}^{(i)} \mathbb{I}\left(\tilde{y}^{(i)}==j\right)log\left(\hat{y}^{(i)}_j\right) \right] \\ 
    & =\sum_{k=1}^{g} \sum_{i \in \overline{A} \cap G_k} \frac{1}{1-\hat{r}_k} \sum_{j=1}^{c}(1-\hat{r}_k)\mathbb{I}\left(y^{(i)}==j\right)log\left(\hat{y}^{(i)}_j\right)  \\ 
    & =\sum_{k=1}^{g} \sum_{i \in \overline{A} \cap G_k} \sum_{j=1}^{c}\mathbb{I}\left(y^{(i)}==j\right)log\left(\hat{y}^{(i)}_j\right)  
\end{align*}
As a reminder, each group $G_k$ is then associated with estimated noise rate $\hat{r}_k=\frac{1}{\vert g_k \vert} \sum_{i \in G_k} 1-\hat{\beta}^{(i)}_{\phi}$ and estimated clean (i.e., correct) rate $1 - \hat{r}_k = \frac{1}{\vert G_k \vert} \sum_{i \in G_k} \hat{\beta}^{(i)}_{\phi}$. We can express the noise and clean rates in terms of $ \hat{\beta}^{(i)}_{\phi}$ since 
\begin{align*}
    & 1 - r_k = \frac{1}{\vert G_k \vert} \sum_{i \in G_k} \mathbb{I}\left(\tilde{y}^{(i)}==y^{(i)}\right) \\
    & = P(y==\tilde{y} \vert \tilde{y}, \textbf{x}) \ for \ a \ random \ instance \ in \ G_k \\
    & = \frac{1}{\vert G_k \vert} \sum_{i \in G_k} P(y^{(i)}==\tilde{y}^{(i)} \vert \tilde{y}^{(i)}, \textbf{x}^{(i)})
\end{align*}
where $r_k$ and $1 - r_k$ are the actual noise and clean rates within group $k$, respectively. Therefore, since $\hat{\beta}_{\phi}$ is trained to predict $P(y==\tilde{y} \vert \tilde{y}, \textbf{x})$, we estimate the noise and clean rates using $\hat{\beta}_{\phi}$.

\section{Preprocessing Details} \label{apd:preprocess}

Here, we provide more detail on our synthetic data generation process and real dataset pre-processing.

\subsection{Synthetic}

Our data generation process is as described below. Note that the $Percentile(p, \{z\})$ function outputs the $p^{th}$ percentile over all values in $\{z\}$. We defined the feature at index 0 to be a synthetic sensitive attribute. Instances with values below the $20^{th}$ percentile for this feature were considered as the `minority', and the rest were considered as the `majority'. Features 10-19 for the majority instances and features 20-29 for the minority instances were set to 0 to provide more contrast between the two groups. For individual $i$,
\begin{align*}
    &d=30, \textbf{x}^{(i)} \sim N(0, 1)^{30} \\
    &\textbf{w} \sim N(0, 1)^{30}, z^{(i)}=\textbf{x}^{(i)} \cdot \textbf{w} \\
    &y^{(i)}=1 \ if z^{(i)}>Percentile(50, \{z^{(j)}\}_{j=1}^{5000}) \ else \ 0 \\
    &x^{(i)}_{j}=0 \ for \j=10,11,...,19\ \\
    &\hspace{8ex} if \ x^{(i)}_0 >Percentile(20, \{x^{(j)}_0\}_{j=1}^{5000}) \\
    &x^{(i)}_{j}=0 \ for \j=20,21,...,29\ \\
    &\hspace{8ex} if \ x^{(i)}_0 <Percentile(20, \{x^{(j)}_0\}_{j=1}^{5000}) 
\end{align*}

\subsection{MIMIC-III}

Data were processed using the FlexIble Data Driven pipeLinE (FIDDLE), [\citep{tang2020democratizing}], a publicly available pre-processing tool for electronic health record data. We used the same features as [\citep{tang2020democratizing}] for our tasks. More information can be found at https://physionet.org/content/mimic-eicu-fiddle-feature/1.0.0/.

\subsection{Adult}

Although, we used a pre-processed version of this dataset, we omitted features pertaining to education, work type, and work sector to make the task more difficult. More specifically, in the file `headers.txt' at the repository mentioned in Footnote 1, we kept all features beginning with `age', `workclass', `education', `marital status', and `occupation'. We also kept the `Sex\_Female' feature. The remaining features were excluded to make the task more difficult. Values were normalized for each feature to have a range of 0-1 by subtracting by the minimum value observed among all individuals and dividing by the range. During training, we only used 1,000 randomly selected individuals from the provided dataset to make the task more difficult, since there would be fewer samples from which to learn. We made the task more difficult for this dataset to further highlight the differences in performance between the approaches.

\subsection{COMPAS}
Although, we used a pre-processed version of this dataset, we omitted the feature `score\_factor' (i.e., the risk score for recidivism from the ProPublica model) to make the task more difficult.  Values were normalized for each feature to have a range of 0-1 by subtracting by the minimum value observed among all individuals and dividing by the range.

\section{Additional Network and Training Details} \label{apd:net_and_train}

Here, our ranges of hyperparameters and implementation choices for the proposed network. All networks were trained on Intel(R) Xeon(R) CPUs, E7-4850 v3 @ 2.20GHz and Nvidia GeForce GTX 1080 GPUs. All layers were initialized with He initialization from a uniform distribution.  We divide our training data into five batches during training. All random seeds (for Pytorch, numpy, and Python's random) were initialized with 123456789.

\begin{table*}[h]
\floatconts
{tab:hyperparams}
{\caption{For each dataset, we list the range of hyperparameters considered for each dataset. For each hyperparameter, the lower bound is shown in the top row, and the upper bound is shown in the bottom row. For hyperparameters we did not tune, only one row is shown.}}
{
    \begin{tabular}{|c|c|c|c|c|c|}
    \hline
     Hyperparameter & Synthetic & MIMIC-ARF & MIMIC-Shock & Adult & COMPAS  \\
     
    \hline \hline
    Layer Size & 10 & 500 & 500 & 100 & 10 \\
    \hline \hline
    Learning Rate & 0.00001 & 0.00001 & 0.000001 & 0.00001 & 0.0001 \\
    \cline{2-6}
    & 0.01 & 0.001 & 0.001 & 0.01 & 0.05 \\
    \hline \hline
    L2 Constant & 0.0001 & 0.000001 & 0.0001 & 0.0001 & 0.0001 \\
    \cline{2-6}
    &  0.1 & 0.01 & 0.1 & 0.1 & 0.01 \\
    \hline \hline
    Filter Threshold & 0.40 & 0.50 & 0.50 & 0.50 & 0.50 \\
    \cline{2-6}
    & 1.00 & 1.00 & 1.00 & 0.90 & 0.90 \\
    \hline \hline
    Noise Added & 0.00001 & 0.00001 & 0.00001 & 0.0001 & 0.0001 \\
    \cline{2-6}
     & 0.01  & 0.001 & 0.001 & 0.001 & 0.01 \\
     \hline \hline
    Number of Parts & 1 & 1 & 1 & 1 & 1 \\
    \cline{2-6}
     & 10  & 10 & 10 & 10 & 10 \\
      \hline \hline
    $\alpha_{GPL}$ & 0.01 & 0.1 & 0.001 & 0.01 & 0.01 \\
    \cline{2-6}
     & 1.0  & 1.0 & 1.0 & 1.0 & 1.0 \\
      \hline \hline
    $\alpha_{1Proposed}$ & 0.1 & 0.1 & 0.01 & 0.1 & 0.01 \\
    \cline{2-6}
     & 10.0  & 10.0 & 10.0 & 10.0 & 10.0 \\
     \hline \hline
     $\gamma_{Proposed}$ & 0.1 & 0.1 & 0.01 & 0.1 & 0.01 \\
    \cline{2-6}
     & 10.0  & 10.0 & 10.0 & 10.0 & 10.0 \\
     \hline \hline
     $\alpha_{2Proposed}$ & 0.1 & 0.1 & 0.01 & 0.1 & 0.01 \\
    \cline{2-6}
     & 10.0  & 10.0 & 10.0 & 10.0 & 10.0 \\
     \hline
    \end{tabular}
}
\end{table*}

\subsection{Hyperparameter Values Considered} 

Here, we show the range of values we considered for our random search. More details are provided in \textbf{Table} \textbf{\ref{tab:hyperparams}}. For any hyperparameters associated with the Adam optimizer not mentioned above, we used the default values. Not all hyperparameters were used with each approach. `Filter Threshold' and `Noise Added' were only used with the baseline SLN + Filter. Here, Filter Threshold refers to the minimum value of the predicted probability of the observed label for an instance to be considered `correctly labeled'. For example, if Filter Threshold=0.5, then all examples whose predicted probability for the observed label is at least 0.5 are considered `correct' and used during training. `Number of Parts' was only used with the baseline Transition. `$\alpha_{GPL}$' was only used with the baseline Fair GPL. `$\alpha_{1Proposed}$', `$\alpha_{2Proposed}$', and `$\gamma_{Proposed}$' was only used with the proposed method. Here, `$\alpha_{1Proposed}$' and `$\alpha_{2Proposed}$' correspond to the terms $\alpha_1$ and $\alpha_2$ that were used in the objective functions. We refer to them with the added term `Proposed' in the subscript in this section to distinguish it from the $\alpha$ value used by the baseline Fair GPL.

\subsection{Network Details} \label{app:arch_details}

For the overall architecture, we used a feed forward network with two hidden layers. The auxiliary $\beta$ prediction component was also implemented with two feed forward layers. All layer sizes are as described in \textbf{Table} \textbf{\ref{tab:hyperparams}}. In addition, we used the ReLU activation function. The complete implementation can be found in the attached code.

\section{Expanded Results} \label{apd:extra_exp}

Here, we describe additional results that were not included in the main text. We begin with followup experiments on the synthetic data and then describes results from the real data.

\subsection{Robustness to Noise Rate Expanded}
Here we include the AUROC and AUEOC plotted separately for the experiments where we varied the overall noise rate  and noise disparity.

\begin{figure}[ht]
    \floatconts
    {fig:res_noise_exp_ext}
    {\caption{Robustness to overall noise rate: breakdown of (a) discriminative performance and (b) bias mitigation. Mean and standard deviation for 10 random seeds.}}
    {
        \subfigure[Discriminative Performance.]{\label{fig:res_noise_rate_auroc}
             \includegraphics[width=0.95\linewidth]{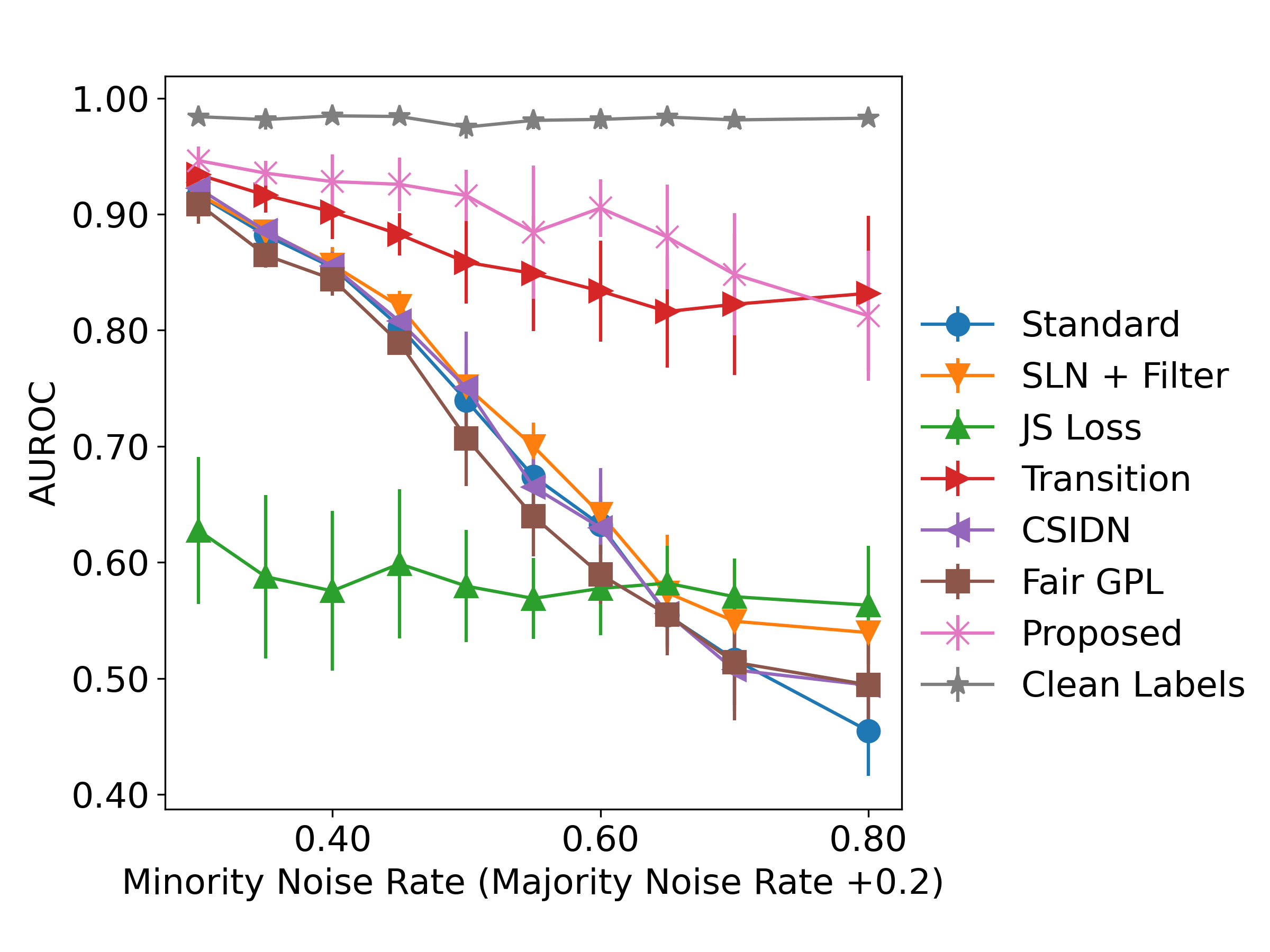} 
            } 
         \qquad       
         \subfigure[Bias Mitigation]{\label{fig:res_noise_rate_aueoc}
             \includegraphics[width=0.95\linewidth]{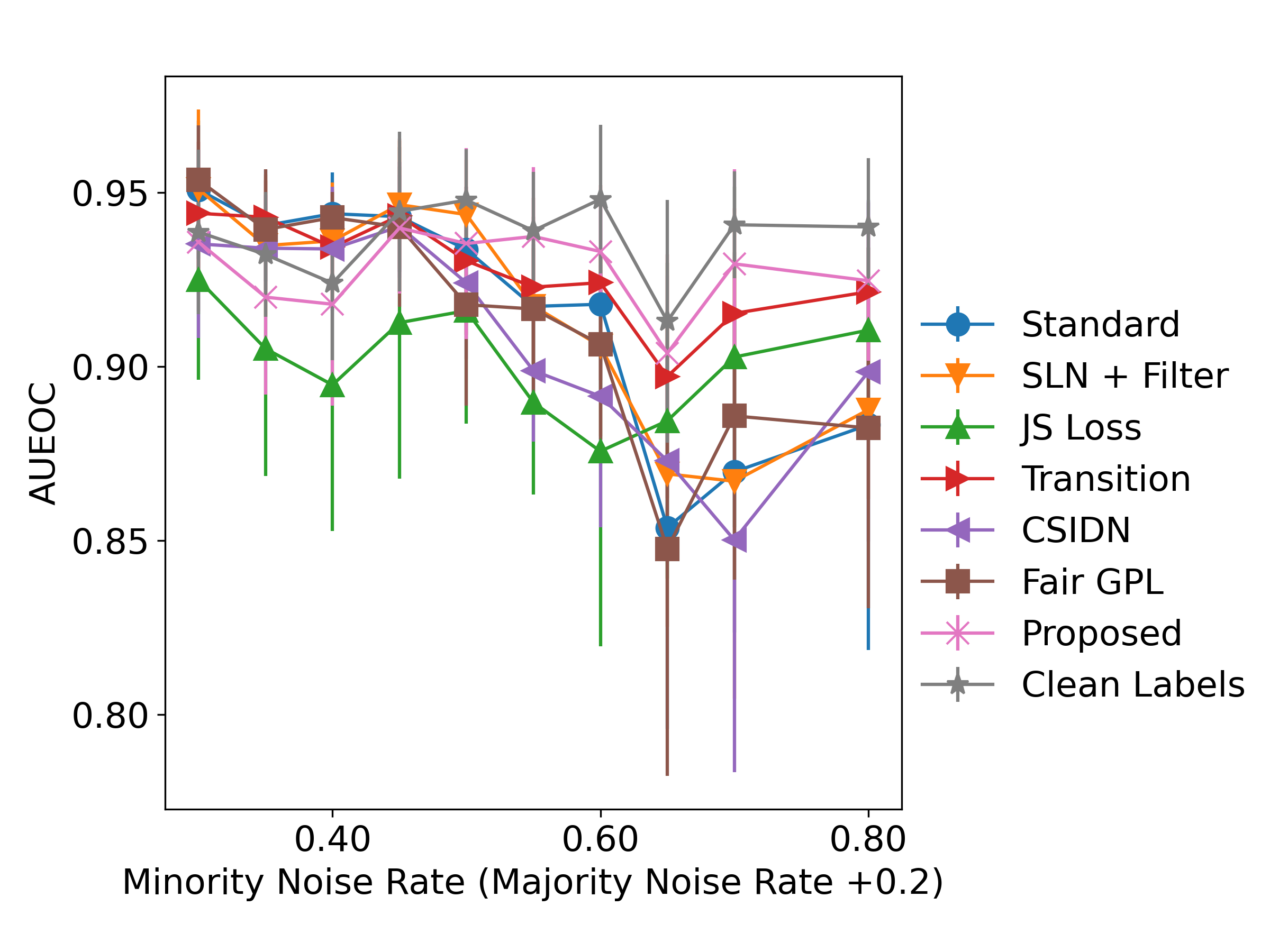}  
            }
     }
\end{figure}

\begin{figure}[ht]
    \floatconts
    {fig:res_noise_disp_ext}
    {\caption{Robustness to the noise disparity: breakdown of (a) discriminative performance and (b) bias mitigation. Mean and standard deviation for 10 random seeds.}}
    {
        \subfigure[Discriminative Performance.]{\label{fig:res_noise_disp_auroc}
             \includegraphics[width=0.95\linewidth]{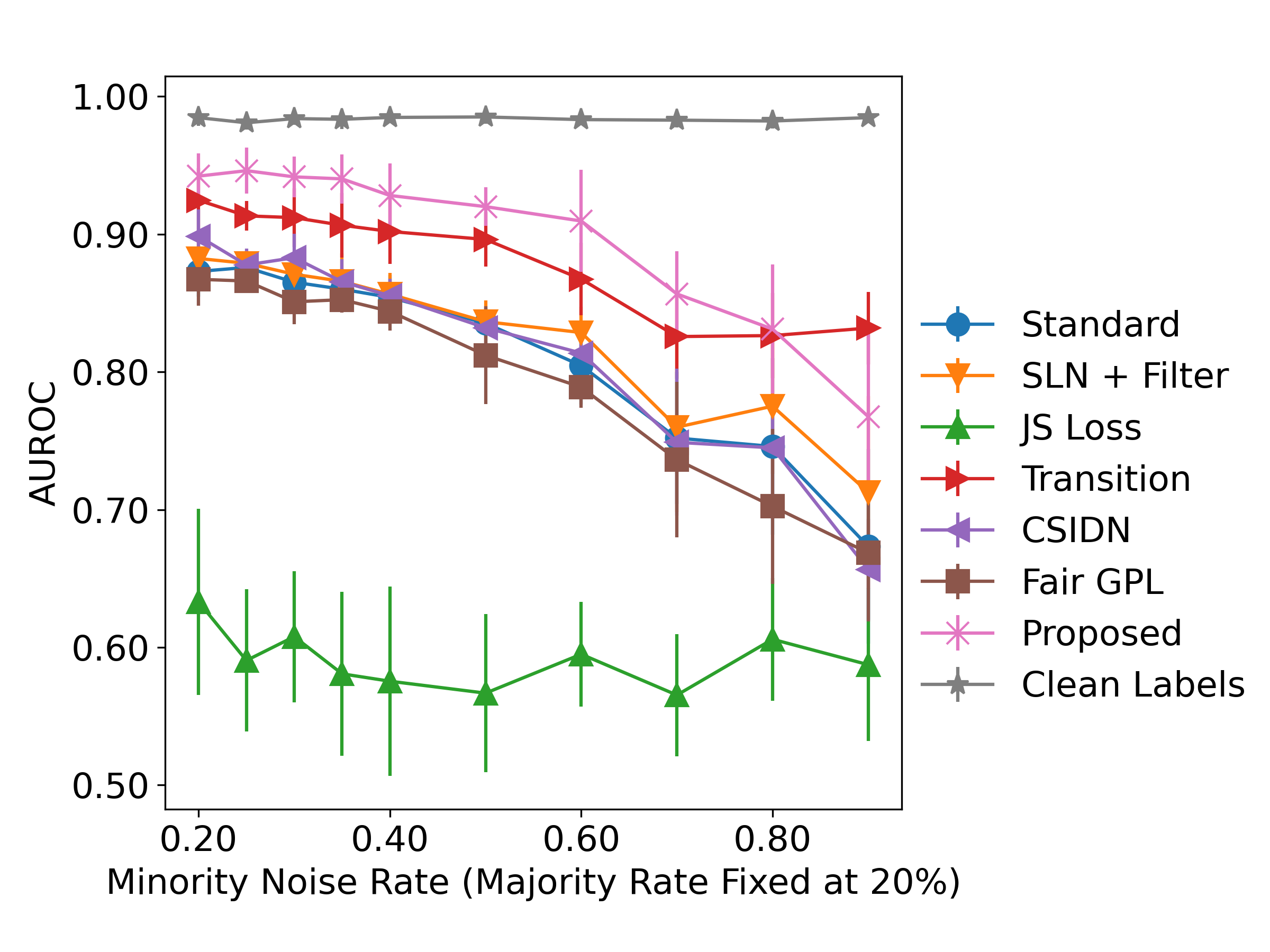} 
            } 
         \qquad       
         \subfigure[Bias Mitigation]{\label{fig:res_noise_disp_aueoc}
             \includegraphics[width=0.95\linewidth]{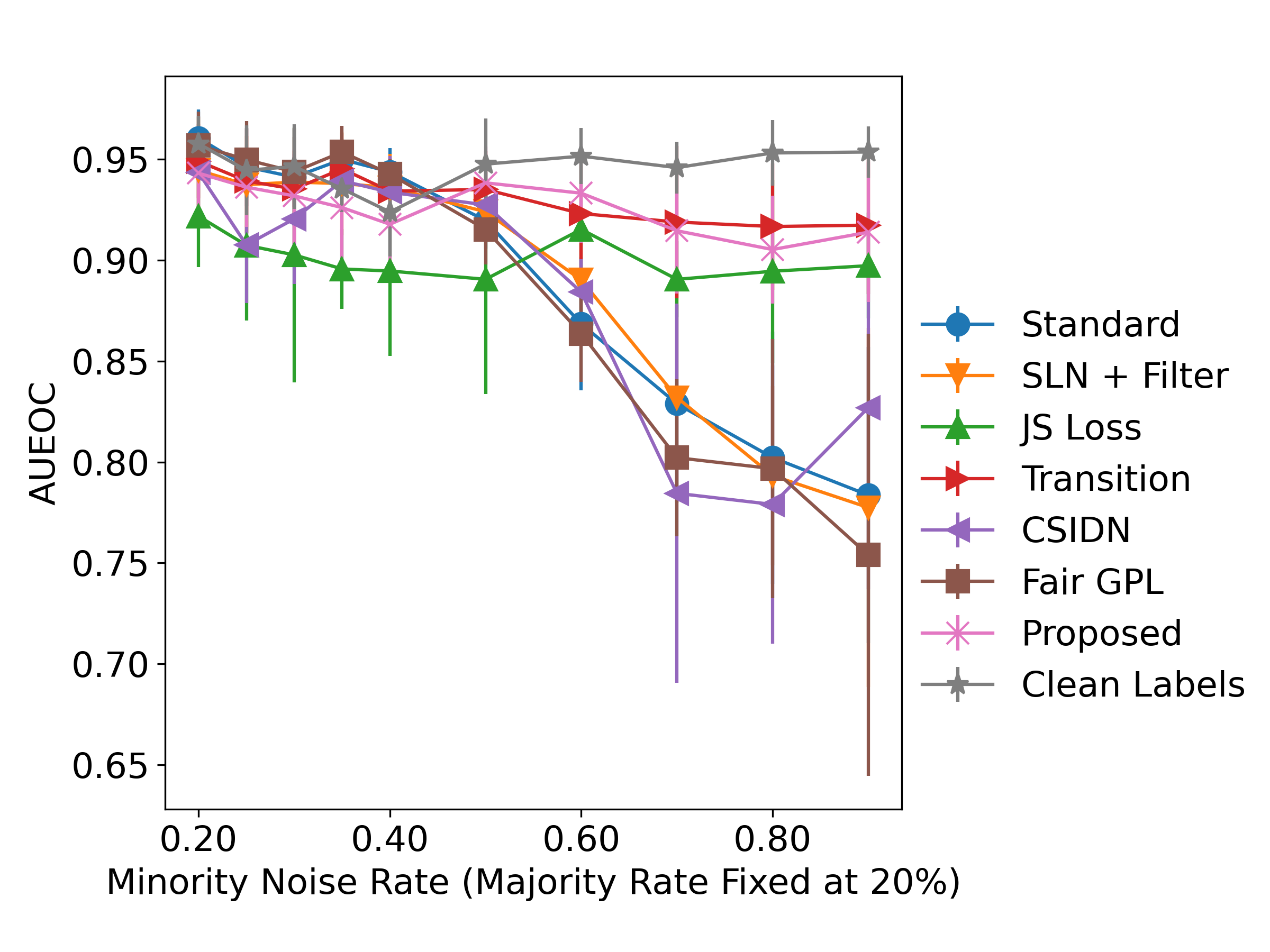}  
            }
     }
\end{figure}

As we varied the overall noise rate (\textbf{Figure} \ref{fig:res_noise_exp_ext}), the proposed approach is able to consistently outperform the baselines with respect to discriminative performance until a minority noise rate of 80\%. This observation is similar to what we observed with the HM. With respect to bias mitigation, the proposed approach is not more beneficial than the baselines up to a minority noise rate of 60\%. At a minority noise rate above 60\%, our approach experienced the least degradation compared to the baseline approaches. This is in line with our expectations since our approach explicitly accounts for differences in noise rates among groups during training.

As we varied the noise disparity (\textbf{Figure} \ref{fig:res_noise_disp_ext}), we have similar observations to the previous experiment in that the proposed approach is able to consistently outperform the baselines with respect to discriminative performance until a minority noise rate of 80\%. With respect to bias mitigation, the proposed approach is not more beneficial than the baselines up to a minority noise rate of 40\%. At a minority noise rate above 40\%, our approach experienced the least degradation compared to most of the other baseline approaches and was comparable to the Transition baseline. Unlike the previous experiment, the degradation in AUEOC among many of the baseline approaches is larger, which is in line with our expectations since we were directly changing the difference in noise rates between the groups while the previous experiment kept the difference constant.

\begin{figure}[ht]
    \floatconts
    {fig:proposed_abl}
    {\caption{Ablation study of proposed approach.}}
    {\includegraphics[width=0.95\linewidth]{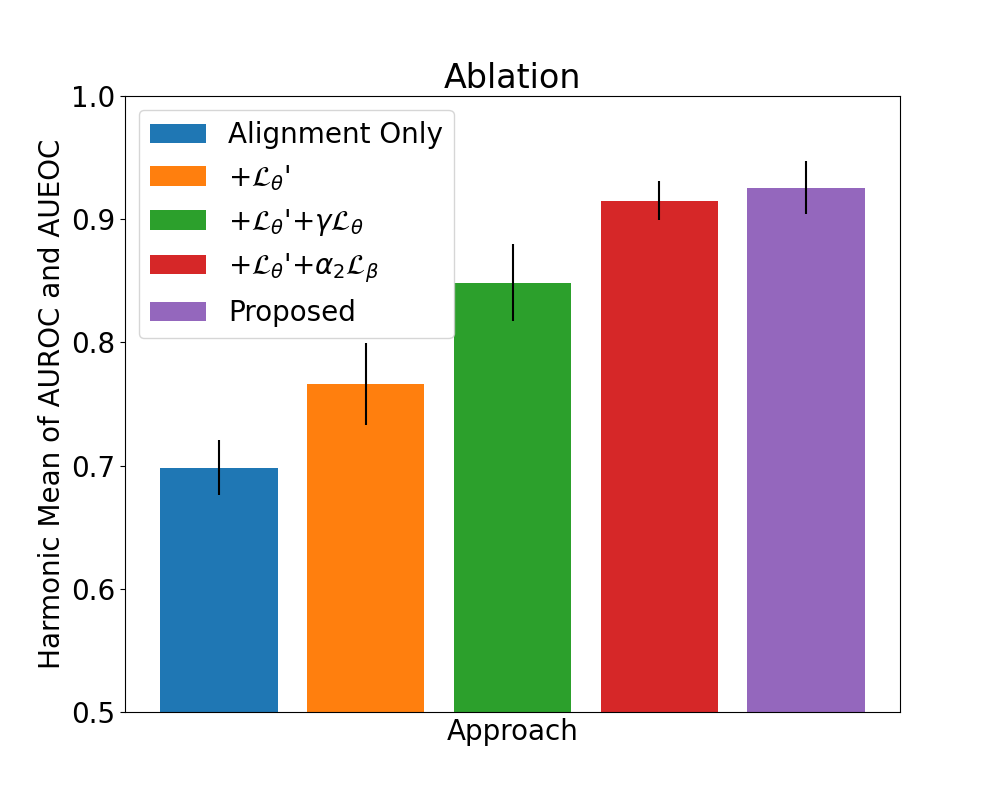}}
\end{figure}

\begin{figure*}[ht]
    \floatconts
    {fig:proposed_sens_analysis}
    {\caption{Sensitivity analysis of proposed approach on objective function hyperparameters.}}
    {
        \subfigure[We varied $\alpha_1$.]{\label{fig:res_alpha1}
             \includegraphics[width=0.31\linewidth]{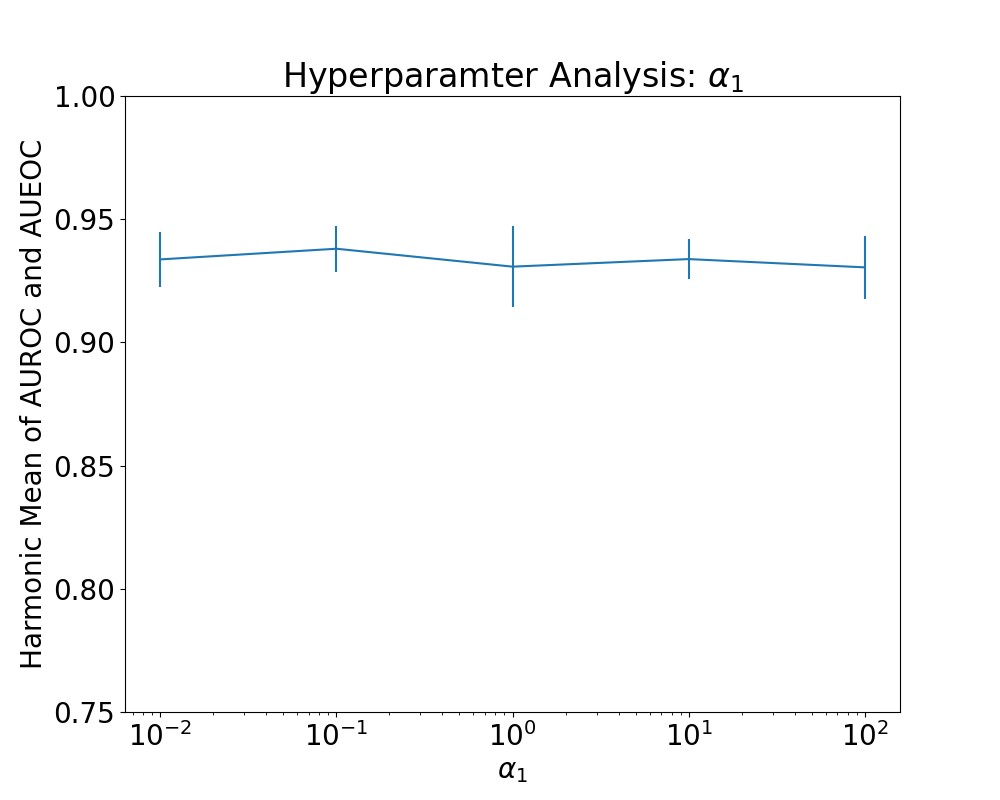} 
            } 
         \subfigure[We varied $\alpha_2$.]{\label{fig:res_alpha2}
             \includegraphics[width=0.31\linewidth]{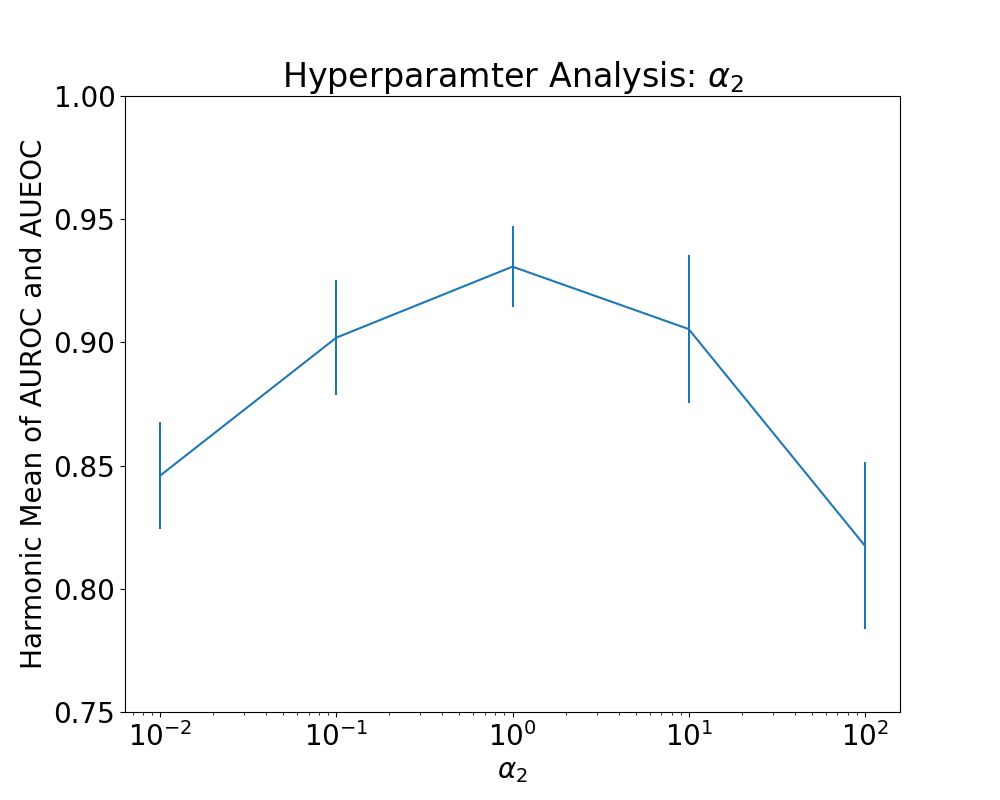} 
            } 
         \subfigure[We varied $\gamma$.]{\label{fig:res_gamma}
             \includegraphics[width=0.31\linewidth]{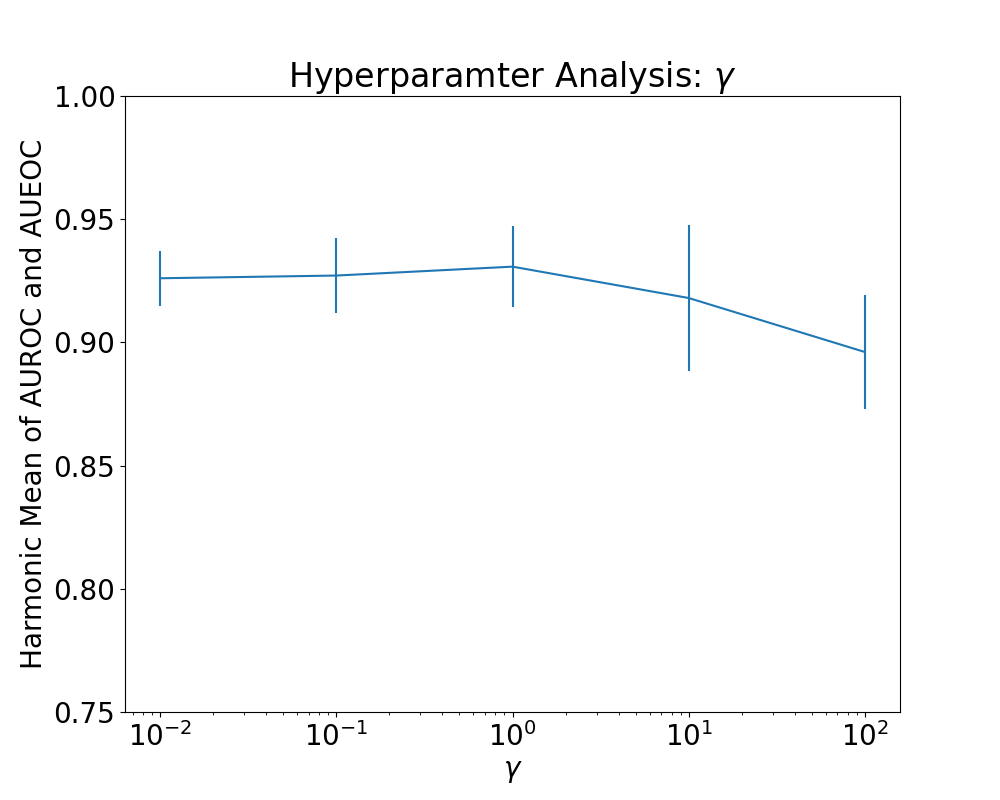} 
            } 
     }
\end{figure*}

\subsection{Ablation Study}
We also examined our approach more closely by conducting an ablation study and a hyperparameter sensitivity analysis on the synthetic data. We used the synthetic dataset since our noise was synthetically introduced and not dataset specific. In our ablation study (\textbf{Figure} \textbf{\ref{fig:proposed_abl}}), we began with training on only the \anchor points (i.e., Step 1 only), which achieved the worst performance. We then introduced Step 2 and added the remaining training data (i.e., non-\anchor points) but only trained using $\mathcal{L}_{\theta'}$. This led to an improvement in performance, but not to the level of the full approach. The next two ablations build on the previous one. In the first one, we added continued supervision on the \anchor points with $\mathcal{L}_{\theta}$, and observed an improvement in performance, likely due to the retention of high quality data in this step. In the second one, we added continued supervision on the \anchor points using $\mathcal{L}_{\phi}$, and observed an even larger improvement. This is likely because including $\mathcal{L}_{\theta}$ prevented the model from learning a solution where $\hat{\beta}$ was small for all instances, as previously discussed. Finally, we end with our full proposed approach, which performed noticeably better than each of the ablations, showing the importance of each component.

\subsection{Hyperparameter Sensitivity Analysis}
In our sensitivity analysis on the synthetic data (\textbf{Figure} \textbf{\ref{fig:proposed_sens_analysis}}), we tested how performance of the (full) proposed approach varied to changes in the hyperparameters $\alpha_1$, $\alpha_2$, and $\gamma$. For each of these hyperparameters, we measured performance at values between 0.01 and 100 on a logarithmic scale while keeping the other two values constant at 1. We found that $\alpha_1$ and $\gamma$ were the most robust to changes in the value. We found that $\alpha_2$ was more sensitive, with values between 0.1 and 10 generally working best. 

\subsection{Sensitivity to \Anchor Set Composition Expanded}
In our analysis on sensitivity to \anchor set composition, we include results for the other baselines in (\textbf{Figure} \textbf{\ref{fig:res_anchor_exp2}}). At \anchor set sizes of below 5\% on the real datasets, the proposed approach was beneficial to the baselines. At larger \anchor set sizes, the baseline Transition was able to match the proposed method due to the increased amount of clean data. When the \anchor set was biased, the proposed approach outperformed the baselines in the unbiased settings and was competitive as bias in the \anchor set increased.

\begin{figure*}[ht]
    \floatconts
    {fig:res_anchor_exp2}
    {\caption{Robustness to varying \anchor sets. Mean and standard deviation for 10 random seeds.}}
    {
        \subfigure[As we decrease the \anchor set size (proportion of training data) performance decreases. Still, at an \anchor set size of 5\%, the proposed approach generally outperforms the baselines.]{\label{fig:res_asize2}
             \includegraphics[width=1\linewidth]{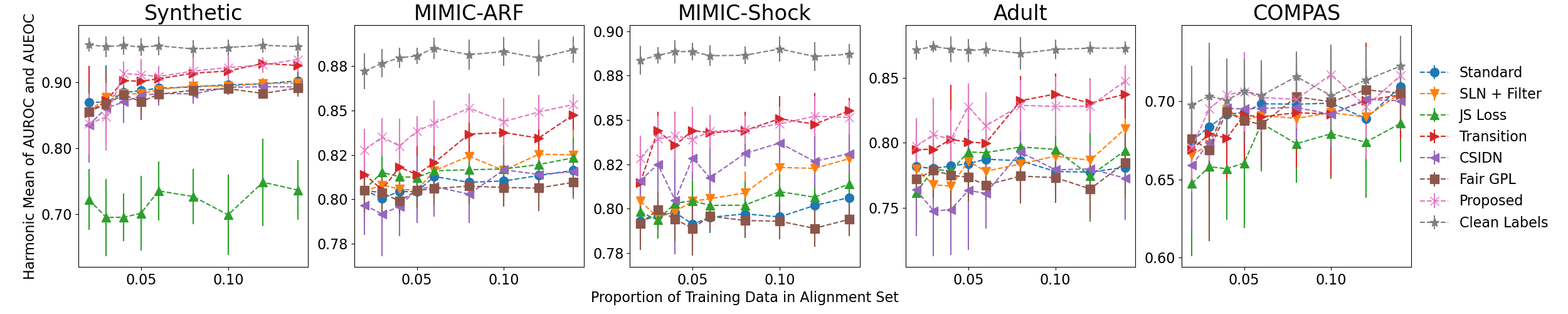} 
            } 
         \qquad       
         \subfigure[As we vary the \anchor set bias (proportion of minority instances) performance varies. The proposed approach is generally robust to changes in the bias of the \anchor set. The dashed vertical black line shows the proportion at which the minority group occurs in the dataset (i.e., an unbiased \anchor set).]{\label{fig:res_adistr2}
             \includegraphics[width=1\linewidth]{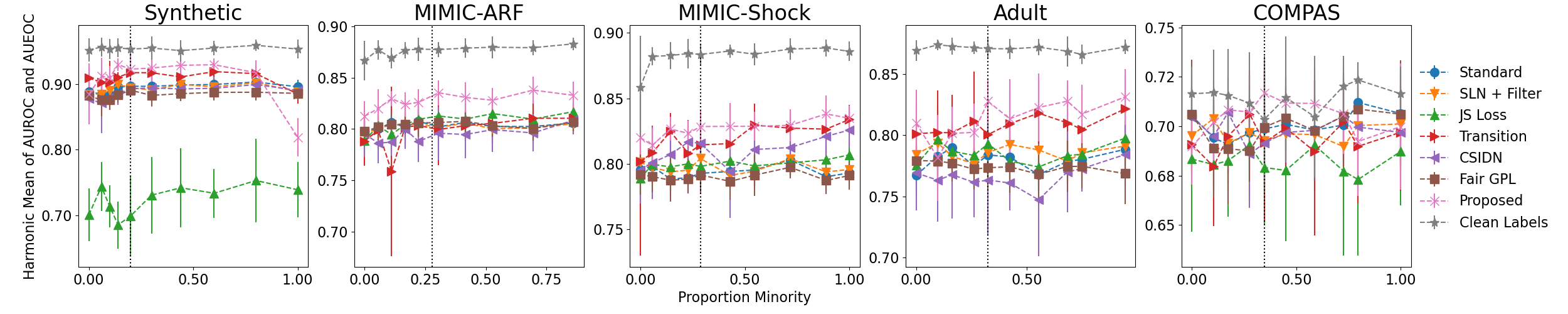} 
            }
     }
\end{figure*}

\end{document}